\documentclass[letterpaper, 10 pt, conference]{ieeeconf}  %

\IEEEoverridecommandlockouts                              %

\usepackage{graphicx} %
\usepackage{amsmath} %
\usepackage{amssymb}  %
\usepackage{cite}
\usepackage{comment}
\usepackage{subcaption}
\usepackage{hyperref}
\usepackage{arydshln}
\usepackage{xcolor}
\usepackage{multirow}
\usepackage{booktabs}       %

\newcommand{\TODO}[1]{}

\title{\LARGE \bf Train Offline, Test Online: A Real Robot Learning Benchmark}

\author{Gaoyue Zhou*$^{1}$, Victoria Dean*$^{1}$, Mohan Kumar Srirama$^{1}$, Aravind Rajeswaran$^{2,5, 6}$, Jyothish Pari$^{3}$, \\
Kyle Hatch$^{4}$, Aryan Jain$^{5}$, Tianhe Yu$^{4}$,
Pieter Abbeel$^{5}$, Lerrel Pinto$^{3}$, Chelsea Finn$^{4}$, Abhinav Gupta$^{1}$
\thanks{*This work was supported by Schmidt Futures, NSF GRFP, and Honda.}%
\thanks{$^{1}$Carnegie Mellon University Robotics Institute}
\thanks{$^{2}$Meta AI}
\thanks{$^{3}$New York University}
\thanks{$^{4}$Stanford University}
\thanks{$^{5}$University of California, Berkeley}
\thanks{$^{6}$University of Washington}
}
\begin{document}

\maketitle
\thispagestyle{empty}
\pagestyle{empty}

\begin{abstract}
Three challenges limit the progress of robot learning research: robots are expensive (few labs can participate), everyone uses different robots (findings do not generalize across labs), and we lack internet-scale robotics data. We take on these challenges via a new benchmark: Train Offline, Test Online (TOTO). TOTO provides remote users with access to shared robotic hardware for evaluating methods on common tasks and an open-source dataset of these tasks for offline training. Its manipulation task suite requires challenging generalization to unseen objects, positions, and lighting. We present initial results on TOTO comparing five pretrained visual representations and four offline policy learning baselines, remotely contributed by five institutions. The real promise of TOTO, however, lies in the future: we release the benchmark for additional submissions from any user, enabling easy, direct comparison to several methods without the need to obtain hardware or collect data.
\end{abstract}

\section{INTRODUCTION}
One of the biggest drivers of success in machine learning research is arguably the availability of benchmarks. From GLUE \cite{wang2018glue} in natural language processing to ImageNet \cite{deng2009imagenet} in computer vision, benchmarks have helped identify fundamental advances %
in many areas. Meanwhile, robotics struggles to establish common benchmarks due to the physical nature of evaluation. The experimental conditions, objects of interest, and hardware vary across labs, often making methods sensitive to implementation details. %
Finally, the difficulties of purchasing, building, and installing %
infrastructure %
make it challenging for newcomers to contribute to the field.

For robotics research to advance, we clearly need a common way to evaluate and benchmark different algorithms. A good benchmark will not only be fair to all algorithms but also have a low participation barrier: setup to evaluation time should be as low as possible. Efforts like YCB~\cite{calli2015ycb} and the Ranking-Based Robotics Benchmark (RB2)~\cite{dasari2021rb2} have aimed to standardize objects and tasks%
, but the onus of setting up infrastructure still lies with each lab. A simple way to overcome this is the use of a common physical evaluation site, as the Amazon Picking Challenge \cite{correll2016analysis} and DARPA Robotics Challenges \cite{buehler2009darpa,krotkov2017darpa,seetharaman2006unmanned} have done. However, the barrier is still high since participants must set up their own training infrastructure. Both of the above frameworks leave the method development phase unspecified and struggle to provide apples to apples comparisons.
\begin{figure}[thpb]
\centering
\includegraphics[trim=0 0 0 0, clip, width=\linewidth]{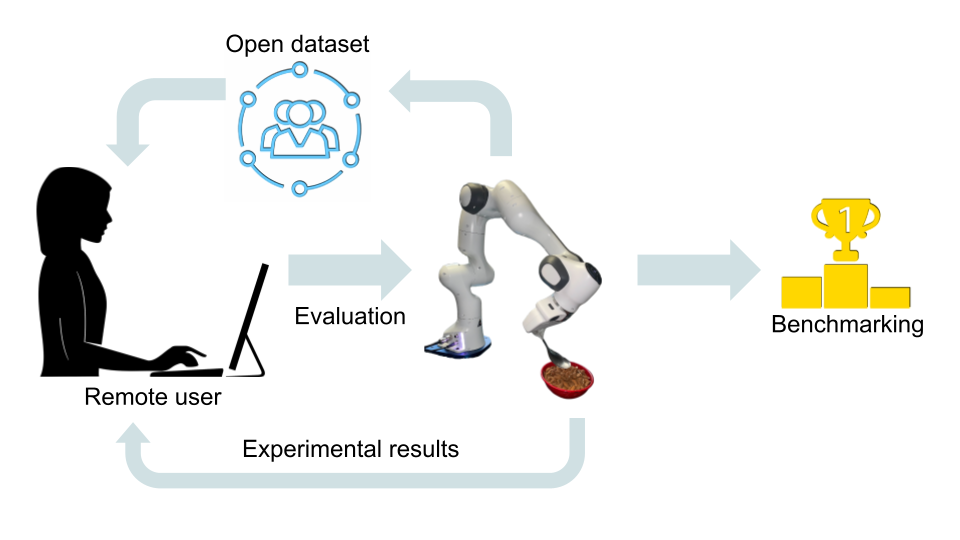}
\caption{\textbf{Train Offline, Test Online}: Our benchmark lets remote users test offline learning methods on shared hardware.}
\label{fig:teaser}
\vspace{-4mm}
\end{figure}

Many robot learning algorithms do online training, where a policy is learned concurrently with data collection. %
One way to standardize online training is with simulation  \cite{todorov2012mujoco,yu2020meta,brockman2016openai,zhu2020robosuite}. While simulation mitigates issues with variation across labs, the findings from simulated benchmarks may not transfer to the real world. On the other hand, if we conduct online training in the real world, comparison across labs becomes difficult due to physical differences. %
In recent years, larger offline datasets have surfaced in robotics \cite{dasari2019robonet, mandlekar2018roboturk, collins2019benchmarking}, and with them the rise of offline training algorithms. From imitation learning to offline reinforcement learning (RL), these algorithms can be trained using the same data and tested in a common physical setup. 

Inspired by this observation, we propose a new robotics benchmark called \textbf{TOTO (Train Offline, Test Online)}. TOTO has two key components: (a) an offline manipulation dataset to train imitation learning and offline RL algorithms, and (b) a shared hardware setup where users can evaluate their methods now and going forward. Because all TOTO participants train using the same publicly-released dataset and evaluate on shared hardware, the benchmark provides a fair apples-apples comparison. 

TOTO paves a path forward for robot learning by lowering the entry barrier: when designing a new method, a researcher can train their policy on our dataset, evaluate it on our hardware, and directly compare it to the existing baselines for our benchmark. TOTO means no more time devoted to setting up hardware, collecting data, or tuning baselines for one individual's environment. In this paper, we lay out our benchmark design and present the initial methods contributed by benchmark beta testers across the country. These results show that our benchmark suite is challenging yet possible, providing room for growth as users iterate on TOTO.

\section{RELATED WORK}
For a thorough description of work related to remote robotics benchmarking, we refer to the Robotics Cloud concept paper \cite{dean2022robots}. Here we describe related work specific to our instantiation of a robotics cloud (TOTO).

\subsection{Shared Tasks and Environments}
A necessary step in comparing method performance is evaluation on a common task. Common tasks might mean a standard object set such as YCB \cite{calli2015ycb}, which can be distributed to remote labs, allowing for shared metrics like grasp success on these objects. RB2 \cite{dasari2021rb2} provides four common manipulation tasks (similar to those we use, described in Section \ref{sec:tasks}) as well as a framework for comparing and ranking methods across results from multiple labs. Another route is sharing the environment itself, as the Amazon Picking Challenge \cite{correll2016analysis} and DARPA Robotics Challenges \cite{buehler2009darpa,krotkov2017darpa,seetharaman2006unmanned} have done. Sharing tasks or environments gives metrics by which we can compare approaches. However, users must still develop the approach on their own hardware in their own lab, and recreating identical environment setups is quite challenging.

\subsection{Shared, Remote Robots}
Going one step further, remotely-accessible robots can be shared across the community, enabling method development and evaluation without users acquiring their own hardware. Georgia Tech's Robotarium \cite{pickem2017robotarium} allows for remote experimentation of multi-agent methods on a physical robotic swarm, which has been extensively used not just in research but also in education. OffWorld Gym \cite{kumar2019offworld} provides remote access to navigation tasks using a mobile robot with closely mirrored simulated and physical instances of the same environment. A recent survey paper \cite{sun2021research} provides an overview of robotic grasping and manipulation competitions, including some involving remotely-accessible, shared robots such as \cite{liu2021ocrtoc}. Finally, most closely related to our work, the Real Robot Challenge \cite{funk2021benchmarking} runs a tri-finger manipulation competition on cube reorientation tasks. The success of the Real Robot Challenge inspires our work, which also allows for evaluation of manipulation tasks on shared robots. Our work, however, is designed to evaluate generalization in robot learning through challenging variations (lighting, unseen test objects, etc.) and an image-based dataset (as opposed to assuming ground-truth state access).

\subsection{Open-Source Robotics Datasets}
Collecting real-world robotics data is challenging and expensive due to physical constraints like environment resets and hardware failures. Thus open-source robotics datasets serve an important role in the field by enabling larger-scale offline robot learning. Some work has improved the way we collect robotics data, such as self-supervised grasping \cite{pinto2016supersizing} and further parallelization of robots \cite{levine2018learning}. RoboTurk \cite{mandlekar2018roboturk} provides a system for simple teleoperated data collection which can be executed remotely. Much work in robot learning has introduced datasets more generally, such as MIME \cite{sharma2018multiple} (8260 demonstrations over 20 tasks), RoboNet \cite{dasari2019robonet} (162,000 trajectories collected across 7 robots), and Bridge Data (7,200 demonstrations across 10 environments). However, it is hard to understand the value of these datasets without a common evaluation platform, something that \cite{collins2019benchmarking} addresses by using simulation to replicate a real-world dataset. In contrast, we address this issue with real-world evaluation that matches the domain of the data collection. Our initial dataset is 2,898 trajectories, but this will grow over time as we add evaluation trajectories collected from users' policies.

\subsection{Offline Robot Learning}
Our benchmark focuses on offline robot learning, including imitation learning and offline RL. Our initial set of baselines is described and contextualized in Section \ref{sec:policy_baselines}.

\section{THE TOTO BENCHMARK} %
Our benchmark focuses on manipulation due to the lack of benchmarking in this area. Our hardware (Section \ref{sec:hardware}) is set in environments that enable a set of benchmark manipulation tasks described in Section \ref{sec:tasks}. We collect an initial dataset on these tasks, detailed in Section \ref{sec:dataset}. Finally, in Section \ref{sec:eval_protocol}, we present the evaluation protocol for all policies contributed to our benchmark. For more information about our dataset and contributing to the benchmark, please see: \url{https://toto-benchmark.org/}.

\subsection{Hardware}
\label{sec:hardware}
Our hardware includes a Franka Emika Panda robot arm and workstation for real-time inference. A simple joint position control stack runs at 30 Hz. The actions are joint targets, which are converted to motor control signals using a high-frequency PD controller. We also provide an end effector controller in which actions are specified via the position and orientation of the gripper. End effector control using X, Y, Z positions alone is not feasible to solve our tasks: for example, the orientation of the gripper must change as the robot pours. 
All the results presented in this paper were attained using the joint position controller. We use an Intel D435 RealSense camera for recording RGB-D image observations.

We allow users to opt for a lower control frequency if desired. The training data can be subsampled by taking one of N frames since the actions are in absolute joint angles. We decrease the test time control frequency accordingly.

\subsection{Tasks}
\label{sec:tasks}
\begin{figure}[thpb]
\centering
\includegraphics[width=\linewidth]{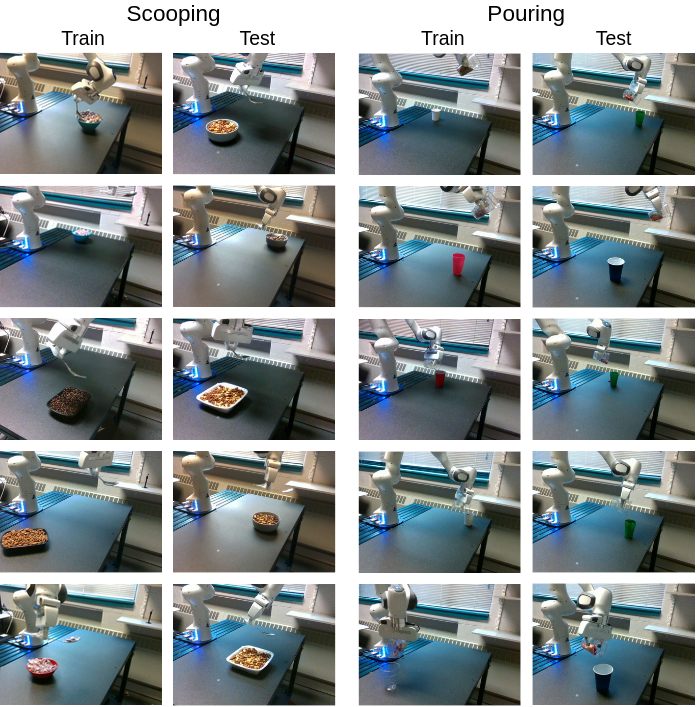}

\caption{\textbf{TOTO Task Suite.} Our benchmark tasks are pouring and scooping, similar to those in RB2 \cite{dasari2021rb2}. Each involves challenging variations in objects, position, and more.}
\label{fig:envs}
\vspace{-5.5mm}
\end{figure}
The task suite consists of two manipulation tasks that humans encounter every day, similar to those introduced in prior work \cite{dasari2021rb2, bahl2021hierarchical}. The tasks are pouring and scooping, excluding the easiest and hardest RB2 tasks (zipping and insertion). Example image observations for these tasks are shown in Fig. \ref{fig:envs}. To see the original task designs, please refer to RB2: \url{https://agi-labs.github.io/rb2/}. Our tasks differ from those in RB2 in a few ways. We randomize the robot start state at the beginning of each episode. We apply a bit more noise to the target object locations. We use different combinations of objects based on availability. Lastly, we do not normalize the reward: the reward is the weight in grams of the material successfully scooped or poured. For detailed information on the task configurations, such as locations and objects, see our website. %

\paragraph{Scooping} The robot starts with a spoon in its gripper and a bowl of material on the table. The objective is to scoop material from the bowl into the spoon. The training set includes all combinations of three target bowls, three materials, and six bowl locations (front left, front center, front right, back left, back center, and back right). %

\paragraph{Pouring} The robot starts with a cup containing granular material in its gripper. The goal is to pour the material into a target cup on the table. The training set includes all combinations of four target cups, two materials, and six target cup locations (same locations as scooping). %
The cup in the robot gripper is always clear plastic, enabling better perception of the material remaining in the cup.

\vspace{-1mm}
\subsection{Dataset}
\label{sec:dataset}
A key pillar of our benchmark is the release of a manipulation dataset. Dataset statistics (number of trials, average trajectory length, success rate, and data collection breakdown) are shown in Table \ref{tab:dataset}. We consider a trajectory successful if it obtains a positive reward, and unsuccessful if the reward is zero. The initial release includes between 1000 and 2000 trajectories per task. Pouring data collection using replay and behavior cloning proved challenging to reset (unsuccessful trials require more cleanup), so it was nearly all collected with teloperation. Each trajectory includes images, robot actions (joint angle targets), joint states (joint angles), and rewards. To improve diversity, the data were collected with three techniques, each described below.
\setlength\tabcolsep{3.5pt} %
\begin{table}[!htbp]
\caption{Dataset overview}
\vspace{-2mm}
\label{tab:dataset}
\begin{center}
\begin{tabular}{lccccccc}
\toprule
&\multicolumn{3}{c}{Task statistics} & \multicolumn{3}{c}{Collection technique} \\
\cmidrule(lr){2-4} \cmidrule(lr){5-7}
& Trials & Length & Success & Teleop & BC & Replay\\ %
\midrule
\texttt{Scooping}         & 1895 & 495 & 0.690 & 41\% & 33\% & 26\%  \\ %
\texttt{Pouring}         & 1003 & 324 & 0.977 & 99\% & \ 0\% & \ 1\%  \\ %
\bottomrule
\end{tabular}
\end{center}
\vspace{-7mm}
\end{table}

\paragraph{Teleoperation}
We collected the majority of trajectories with teleoperation using Puppet \cite{kumar2015mujoco}. The human controls the robot in an intuitive end effector space using an HTC Vive virtual reality headset and controller. While this teleoperation is theoretically possible to use remotely, we collect the data with the human and robot in the same room, giving the human direct perception of the scene. Our multiple teleoperators have different dominant hands, leading to more diverse data. Most teleoperation trials are successful.

\paragraph{Behavior cloning rollouts}
After teleoperation trajectories are collected, we train simple, state-based behavior cloning (BC) policies on each target location, so no visual perception is required. We roll out BC trajectories with some noise added to actions at each timestep. The amount of noise varies across trajectories for additional diversity.

\paragraph{Trajectory replay}
Finally, we also replay individual teleoperated trajectories with added noise. While these might seem overly similar to the original teleoperated trajectories, keep in mind that conditions like lighting also vary with time of day, so this replay still expands the dataset in other ways.

\subsection{Evaluation Protocol}
\label{sec:eval_protocol}
To evaluate each task, we use two unseen objects (bowls and cups) and one unseen material (mixed nuts for scooping and Starburst candies for pouring). We evaluate three object locations seen during training (front left, front center, front right) and three unseen locations. We evaluate three training seeds of each method. We initialize the robot with a randomly sampled pose at the beginning of each trajectory. However, the robot's initial poses are kept the same across seeds to ensure minimal variance. Combining 2 objects, 1 material, 3 locations, and 3 seeds means that each method is evaluated across 18 trials each for train and test locations. We report mean and variance of these trials.%

\section{BENCHMARK USE}
Here we introduce the framework for our benchmark. TOTO is designed to make the user workflow (Section \ref{sec:workflow}) easy for newcomers with well-documented software infrastructure (Section \ref{sec:software_infrastructure}) including examples and tests.

\subsection{User Workflow}
\label{sec:workflow}
We provide a real-world dataset (Section \ref{sec:dataset}) collected using our hardware setup (Section \ref{sec:hardware}). Participants optionally use our software starter kit (Section \ref{sec:software_infrastructure}) and locally train policies of their choosing using this data. Users submit policies through Google Forms for evaluation on our real-world setup. They do not receive the low-level data from these evaluation trials; they simply receive a video showing the policy behavior as well as %
the reward and success rate.%

An engineer supervises the real-world evaluations; thus the evaluation turnaround time is currently around 12 hours (depending on time of day submitted). Our goal is to emphasize offline learning and prevent overfitting, removing the need for real-time results or large quantities of evaluation.

As new users evaluate methods after the paper release, we will post (anonymous) evaluation scores for each attempt on a website leaderboard. We will also periodically add data collected by the users' policies to the original dataset.

\subsection{Software Infrastructure}
\label{sec:software_infrastructure}
Our software starter kit includes documented code and instructions for policy formatting and dataset usage. We have open-sourced baseline code, trajectory data, and pretrained models (see our website). These components ensure that TOTO is easily accessible to a broad portion of the robotics, ML, and even computer vision communities. %

We adapt the agent format from \cite{ke2021grasping}, which requires a \texttt{predict} function taking in the observation and returning the action. We use a standard config format and require an \texttt{init\_agent\_from\_config} function to create the agent. %

We provide users with code for training an example image-based BC agent and a docker environment which wraps the minimum required dependencies to run this code. Users can optionally extend the docker containers with additional dependencies. We also provide a stub environment for users to locally verify whether their agent's predictions are compatible with our robot environment. This setup allows users to resolve all agent format and library dependency issues before submitting agents for evaluation.

\section{BASELINES}
We highlight the importance of establishing a benchmark by running two sets of experiments: (a) what is a good visual representation for manipulation? and (b) what is a good offline algorithm for policy learning?
To test the benchmark infrastructure%
, we have solicited baseline implementations for both experiments from several labs.

\subsection{Visual Representation Baselines}
\label{sec:vision_baselines}
A core unanswered question, due to the lack of benchmarking, is what is a good visual representation for manipulation? Is ResNet trained on ImageNet great or do self-supervised approaches outperform supervised models? We evaluate five visual representations provided by TOTO users from multiple labs. Two are trained on our data (in-domain) and three are generically pretrained.

\texttt{BYOL} (Bootstrap Your Own Latent) \cite{grill2020bootstrap} is a self-supervised representation learning method trained on our dataset. The BYOL representation embedding size is 512.

\texttt{MoCo (Generic)} refers to the Momentum Contrast (MoCo) model trained on ImageNet \cite{he2020momentum}, while \texttt{MoCo (In-Domain)} is trained on our data with crop-only augmentations~\cite{Parisi2022PVR}.

\texttt{Resnet50} refers to the model trained with supervised learning on ImageNet \cite{he2016deep}. %

\texttt{R3M} (Reusable Representations for Robot Manipulation) \cite{nair2022r3m} is trained on Ego4D \cite{grauman2022ego4d} with time-contrastive learning and video-language alignment. For \texttt{R3M}, \texttt{MoCo}, and \texttt{Resnet50}, we use the 2048-dimensional embedding vector following the fifth convolutional layer.

These representations performed the best among a larger set of vision models on which we ran an initial brief analysis (including offline visualizations and BC rollouts). Additional representations that performed less well (and therefore are not included as baselines) included CLIP \cite{radford2021learning} and a lower-level MoCo model (from the third layer instead of the fifth).

\subsection{Policy Learning Baselines}
\label{sec:policy_baselines}
Remote users have contributed the below policy learning baselines, which span the spectrum from nearest neighbor querying to BC to offline RL. They were selected according to each TOTO contributor's expertise with approach coverage in mind. All methods use RGB image observations, and some run these images through a frozen, pretrained vision model before passing the resulting embedding to a policy. %

\texttt{BC} is trained on top of each vision representation baseline. Closed-loop BC %
predicts a new action every timestep, while open-loop BC predicts a sequence of actions to execute without re-planning. Our BC baseline is \emph{quasi} open-loop: training trajectories are split into 50-step action sequences, and the policy is trained to predict such a sequence given the current observation. During evaluation, these 50 actions are executed between each prediction step. We find that this performs better than closed-loop or open-loop alone: closed-loop struggles without history, and open-loop is challenging with our variable-length tasks. We filter the training data to only include trajectories with nonzero reward~\cite{chen2021decision}.

\texttt{VINN} (Visual Imitation through Nearest Neighbors) \cite{pari2021surprising} is a nearest neighbor policy using an image encoder trained with \texttt{BYOL} \cite{grill2020bootstrap}. While using nearest neighbors as a policy has been previously explored \cite{mansimov2018simple}, this approach alone does not scale well to high-dimensional observations like images. BYOL maps the high-dimensional observation space to a low dimension to obtain a robust policy. VINN was originally closed-loop (query and execute a new action at each timestep), but in this work we mirror the 50-step quasi open-loop approach used in the BC baseline (described above).

\texttt{IQL} (Implicit Q-learning) \cite{kostrikov2021offline} uses the open-source implementation from the \texttt{d3rlpy} package \cite{seno2021d3rlpy}. We use \texttt{MoCo (In-Domain)} as a frozen visual representation since it performed the best in our comparison of representations with BC. We concatenate the frozen image embeddings with the robot's joint angles as the input state to the model.

\texttt{DT} (Decision Transformers)~\cite{chen2021decision} recasts offline RL as a conditional sequence modeling task using transformers. Similar to BC, it is trained to predict the action in the dataset, but conditions on the trajectory history as well as target return (desired level of performance). We use the Hugging Face DT implementation.
The model receives an RGB image and the robot's joint angles: the former is embedded using \texttt{MoCo (In-Domain)} and concatenated with the latter at each step.
DT uses a sub-sampling period of 8 and a history window of 10 frames.
For inference and evaluation, the target return prompt is approximately chosen as the mean return from the top 10$\%$ of trajectories in the dataset for each task.

\begin{figure*}[ht]
\centering
\vspace{-2mm}
\includegraphics[trim=0 0 0 0, clip, width=\linewidth]{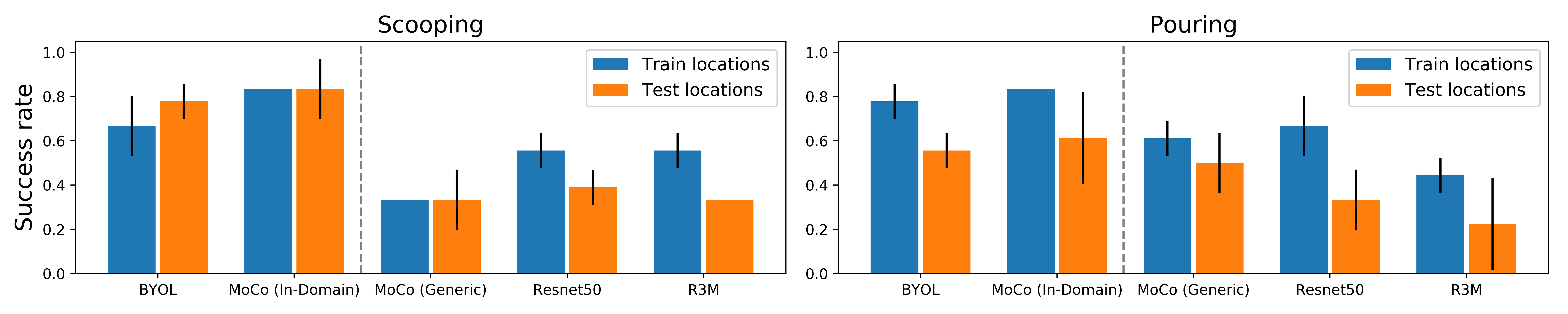} 
\vspace{-4mm}
\caption{\textbf{Vision representation comparison with BC.} Models trained on our data (left of dashed line) perform better than generic ones (right of dashed line), and results tend to be better for training object locations than unseen test locations.}
\vspace{-4mm}
\label{fig:vision_plots}
\end{figure*}

\section{EXPERIMENTAL RESULTS}
\label{sec:result}

\subsection{Visual Representation Comparison Using BC}
\label{sec:visual_representations}
Our first set of experiments compares BC agents using the vision representations detailed in Section \ref{sec:vision_baselines} and evaluated with the protocol described in Section \ref{sec:eval_protocol}. The success rates across all representations and tasks are visualized in Fig. \ref{fig:vision_plots}, and the numerical rewards are presented in Table \ref{tab:vision_results}.

These results show that finetuning the MoCo model on our data outperforms the generic version, as expected. \texttt{MoCo (In-Domain)} achieves the highest success rate and average reward on both tasks, followed by \texttt{BYOL}, the other in-domain model. In general, the relative performance between models is mostly consistent across tasks. Resnet50 and \texttt{MoCo (Generic)} perform slightly better on pouring than on scooping.

Fig. \ref{fig:vision_plots} also visualizes performance differences due to object locations. Locations seen during training perform better, but performance does not degrade significantly, suggesting that the representations have a generalizable notion of where the target object is. Surprisingly, the two representations trained on our data (\texttt{MoCo (In-Domain)} and \texttt{BYOL}) perform equally good or even slightly better on unseen locations for scooping.

\setlength\tabcolsep{4pt} %

\begin{table}[!htbp]
\caption{Performance of vision representations with BC across train and test locations.}%
\label{tab:vision_results}
\vspace{-2mm}
\begin{center}
\begin{tabular}{rlcccc}
\toprule%
  &\multirow{2}[3]{*}{Method}& \multicolumn{2}{c}{Scooping} & \multicolumn{2}{c}{Pouring}  \\
  \cmidrule(lr){3-4} \cmidrule(lr){5-6}
& & \rule{0pt}{2ex} Reward & Success \% & Reward & Success \% \\
\midrule%
\multirow{2}{4.5em}{In Domain}&\texttt{BYOL}      & 4.39 & 72.2\% & 20.22 & 66.6\% \\
&\texttt{MoCo}      & \textbf{7.42} & \textbf{83.3\%} & \textbf{22.86} & \textbf{72.2\%}\\
\midrule%
\multirow{3}{4.5em}{Out of Domain}&\texttt{MoCo}        & 2.11 & 33.3\% & 14.89 & 55.5\%\\
&\texttt{ResNet50}     & 2.83 & 47.2\% & 18.86 & 50.0\% \\
&\texttt{R3M}     & 2.97 & 44.4\% & \ 6.94 & 33.3\%\\
\bottomrule%
\end{tabular}
\end{center}
\vspace{-5mm}
\end{table}

\subsection{Policy Learning Results}
\begin{figure*}[thpb]
\centering
\includegraphics[trim=0 0 0 0, clip, width=\linewidth]{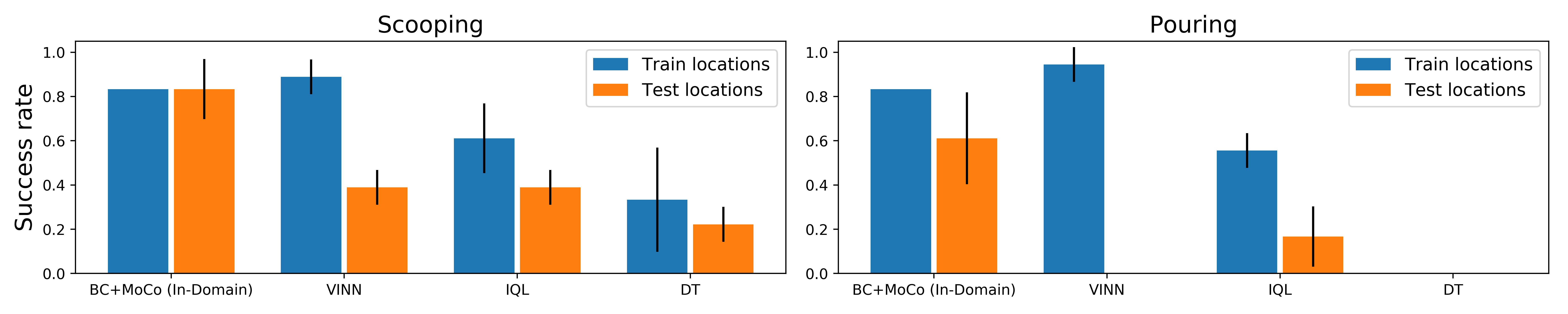}
\vspace{-4mm}
\caption{\textbf{Evaluating offline policy learning results.} \texttt{VINN} sees the best performance on train locations, but its performance degrades on unseen locations, as does the performance of other methods.}
\label{fig:policy_learning_plots}
\vspace{-4mm}
\end{figure*}

Table \ref{tab:policy_results} shows the comparison of policy learning methods (described in \ref{sec:policy_baselines}) evaluated on TOTO. Due to compute constraints, we have 1 and 2 seeds for \texttt{DT} and \texttt{IQL} respectively. We compensate by duplicating the evaluation of these seeds to keep the number of trials consistent. The results are visualized in Fig. \ref{fig:policy_learning_plots}. We find that \texttt{VINN} performs the best in train locations. We also note that offline-RL approaches  (especially IQL) achieve some success unlike in RB2\cite{dasari2021rb2}. This is likely due to a larger and more diverse dataset than RB2, which contributes to better offline RL performance.

In experiments, we found that scooping proves challenging due to a non-markovian aspect of the task: the spoon is above the bowl both before and after scooping. Thus we would expect open-loop methods (\texttt{BC}, \texttt{VINN}) and those with history (\texttt{DT}) to perform better than others in this setting. While \texttt{BC} and \texttt{VINN} achieve competitive performance on scooping, \texttt{DT} only achieves moderate success on scooping and does not see any positive rewards on pouring. Meanwhile, \texttt{IQL} provides decent performance without history on a non-markovian task.

Comparing the train and test location results for policy learning proves interesting. \texttt{VINN} performs the best on train locations, but it struggles on unseen locations, since it selects actions using the nearest neighbor trajectory from the training data. All other methods also experience some level of degradation when moving to unseen locations, leaving one clear direction for method improvement using TOTO.

\begin{table}[!htbp]
\caption{TOTO policy learning results across train and test locations.}
\vspace{-2mm}
\label{tab:policy_results}
\begin{center}
\begin{tabular}{lcccc}
\toprule%
  \multirow{2}[3]{*}{Method}& \multicolumn{2}{c}{Scooping} & \multicolumn{2}{c}{Pouring}  \\
 \cmidrule(lr){2-3} \cmidrule(lr){4-5}%
 & \rule{0pt}{2ex} Reward & Success \% & Reward & Success \% \\
\midrule%

\texttt{BC + MoCo}         & 7.42 & \textbf{83.3\%} & \textbf{22.86} & \textbf{72.2\%} \\
\texttt{VINN}       & \textbf{7.89} & 63.9\% & 21.75 & 47.2\% \\
\texttt{IQL}      & 6.08 & 47.2\% & \ 9.86 & 38.9\% \\
\texttt{DT}     & 2.83 & 27.8\% & \ 0.00 & \ 0.0\% \\
\bottomrule%
\end{tabular}
\end{center}
\vspace{-5mm}
\end{table}

\subsection{Dataset Size Ablation}
To understand the impact of dataset size on policy learning performance, we perform an ablation in which we train BC on the scooping task with varying amounts of data. We sort the scooping trajectories by reward and train policies with the top 5\%, 25\%, 50\%, 100\% of the data, as well as all successful trajectories with positive rewards ($\sim$70\%). This sorting by reward ensures that policies trained in the small-data regime are not overcome by unsuccessful trajectories. We present the dataset size ablation results in Table \ref{tab:size_ablation}. 
\begin{table}[!htbp]
\caption{Dataset size ablation with BC on scooping.} %
\label{tab:size_ablation}
\vspace{-2mm}
\begin{center}
\begin{tabular}{lcc}
\toprule%
 Dataset size & Reward & Success Rate\\
\midrule%
\texttt{5\%}         & 2.89 & 38.9\% \\
\texttt{25\%}         & 5.94 & 72.2\% \\
\texttt{50\%}        & 6.22 & 77.8\% \\
\texttt{Successes ($\sim$70\%)}      & 8.06 & 83.3\% \\
\texttt{100\%}      & 5.00 & 72.2\% \\
\bottomrule%
\end{tabular}
\end{center}
\vspace{-5mm}
\end{table}

The \emph{all success} number uses the same policy as the BC policy in Table \ref{tab:policy_results}, but we evaluate it again with the ablations to ensure minimal variance in conditions. As expected, training on more data generally leads to a higher success rate. Training on all of the data (including unsuccessful trajectories) leads to a lower reward than training on only the successful trajectories, also unsurprising given the use of BC to learn the policies in this ablation (we might expect offline RL to improve with the inclusion of unsuccessful trials).

Overall, these ablation results suggest that the TOTO dataset size is the right order of magnitude in terms of policy learning. We have reached the point of diminishing returns: training on 50\% versus 70\% of the data does not substantially improve performance. However, additional data might still improve visual representation learning.%

\subsection{Metrics for Offline Policy Evaluation}
A TOTO user might wish to sanity check their policy before submitting it for real-world evaluation or otherwise have performance metrics to guide offline tuning. Here we present simple example metrics for offline evaluation: action similarity to a validation set of expert demonstrations using both joint angle error and end effector pose error. From a chosen validation set of 100 trajectories, we estimate the joint angle error and end effector error by computing the mean squared error between agent's predicted actions and actual actions for all samples.

\begin{figure}[thpb]
\centering
\vspace{-1mm}
\includegraphics[trim=0 0 0 0, clip, width=0.95\linewidth]{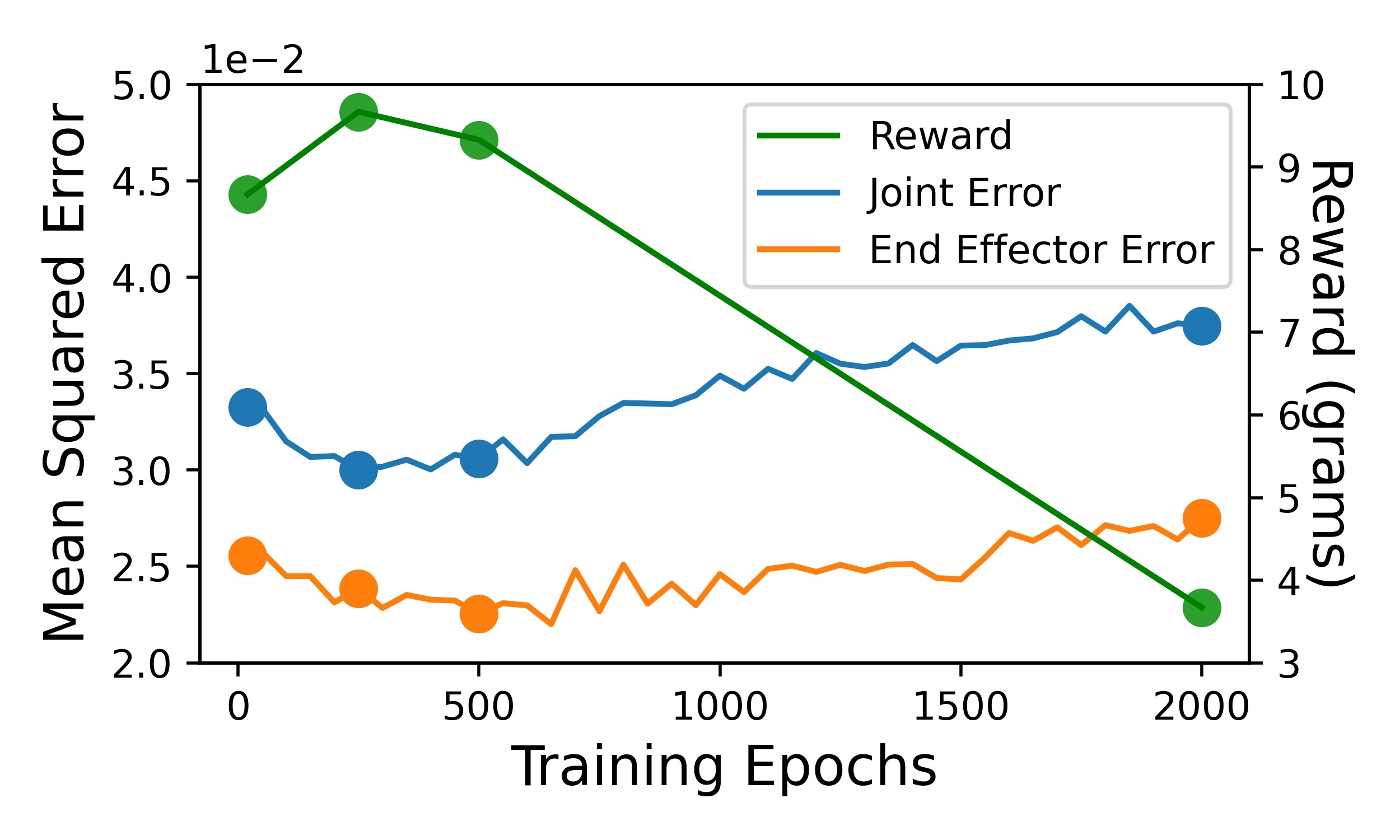}
\vspace{-1mm}
\caption{\textbf{Comparing offline evaluation to online performance.} While offline evaluation is imperfect, it provides a sanity check to the user, guiding development at a higher frequency than real-world evaluation.}
\label{fig:validation_plot}
\vspace{-5mm}
\end{figure}

Fig. \ref{fig:validation_plot} shows these validation metrics on BC checkpoints throughout training and the real-world reward evaluated on four representative checkpoints. The reward increases as the validation error metrics decrease, matching expectations. These metrics capture overfitting: the overtrained policy from 2,000 epochs shows a significant decrease in real-world reward and likewise has higher validation error. While offline metrics alone should not fully guide the development of an algorithm, it provides a signal as to whether the policy might achieve any success in the real world. %

\section{DISCUSSION}
The main goal of this work is to introduce TOTO, our robotics benchmark. We presented a broad initial set of baselines containing both vision representations and policy learning approaches, which can be built off of by future TOTO users. Notably, these baselines were contributed in the same way that TOTO will be used in the future: by collaborators who locally train policies and submit them for remote evaluation on shared hardware. This shows the feasibility of our user workflow. The initial baseline results show the challenging nature of our tasks, especially with respect to generalization. By using TOTO as a community, we can more quickly iterate on ideas and make progress on the real-world bottlenecks to robot learning.

\subsection{Limitations and Future Work} %
The evaluation protocol currently has manual steps: we measure the material transferred during pouring and scooping to compute rewards and reset by returning the material to the original object. We do see future potential to automate reward measurements and resets, such as by adding a scale beneath the target object and using an additional robot to reset the transferred materials. Spills of the transferred material, however, might still require manual intervention.

We plan to expand the evaluation setup to include additional robots. This would help us meet the increasing demand in evaluations as more users adopt the benchmark. One challenge will be visual differences across robots, but we plan to collect additional demonstrations on new robots, and this would be an opportunity to expand the set of tasks as well (we could designate one robot per task).

As user demand further grows, we will implement an evaluation job queue which prioritizes evaluation requests from different users and schedules the jobs based on the number of robots currently available.

\section*{ACKNOWLEDGMENT}
We thank Liyiming Ke for policy training code, Krishna Patel for data collection help, the R-PAD Lab team (Daniel Seita, Sarthak Shetty, Edward Li, and David Held) for beta-testing depth-based methods, Aviral Kumar for offline RL advice, and Gupta Lab beta testers (Sam Powers, Sudeep Dasari, and Himangi Mittal) for starter code feedback.

\bibliographystyle{IEEEtran}
\bibliography{IEEEabrv,papers}

\begin{thebibliography}{10}
\providecommand{\url}[1]{#1}
\csname url@samestyle\endcsname
\providecommand{\newblock}{\relax}
\providecommand{\bibinfo}[2]{#2}
\providecommand{\BIBentrySTDinterwordspacing}{\spaceskip=0pt\relax}
\providecommand{\BIBentryALTinterwordstretchfactor}{4}
\providecommand{\BIBentryALTinterwordspacing}{\spaceskip=\fontdimen2\font plus
\BIBentryALTinterwordstretchfactor\fontdimen3\font minus
  \fontdimen4\font\relax}
\providecommand{\BIBforeignlanguage}[2]{{%
\expandafter\ifx\csname l@#1\endcsname\relax
\typeout{** WARNING: IEEEtran.bst: No hyphenation pattern has been}%
\typeout{** loaded for the language `#1'. Using the pattern for}%
\typeout{** the default language instead.}%
\else
\language=\csname l@#1\endcsname
\fi
#2}}
\providecommand{\BIBdecl}{\relax}
\BIBdecl

\bibitem{wang2018glue}
A.~Wang, A.~Singh, J.~Michael, F.~Hill, O.~Levy, and S.~R. Bowman, ``Glue: A
  multi-task benchmark and analysis platform for natural language
  understanding,'' \emph{arXiv preprint arXiv:1804.07461}, 2018.

\bibitem{deng2009imagenet}
J.~Deng, W.~Dong, R.~Socher, L.-J. Li, K.~Li, and L.~Fei-Fei, ``Imagenet: A
  large-scale hierarchical image database,'' in \emph{Conference on Computer
  Vision and Pattern Recognition}.\hskip 1em plus 0.5em minus 0.4em\relax IEEE,
  2009, pp. 248--255.

\bibitem{calli2015ycb}
B.~Calli, A.~Singh, A.~Walsman, S.~Srinivasa, P.~Abbeel, and A.~M. Dollar,
  ``The ycb object and model set: Towards common benchmarks for manipulation
  research,'' in \emph{International Conference on Advanced Robotics}.\hskip
  1em plus 0.5em minus 0.4em\relax IEEE, 2015, pp. 510--517.

\bibitem{dasari2021rb2}
S.~Dasari, J.~Wang, J.~Hong, S.~Bahl, Y.~Lin, A.~S. Wang, A.~Thankaraj, K.~S.
  Chahal, B.~Calli, S.~Gupta \emph{et~al.}, ``Rb2: Robotic manipulation
  benchmarking with a twist,'' in \emph{NeurIPS Datasets and Benchmarks Track},
  2021.

\bibitem{correll2016analysis}
N.~Correll, K.~E. Bekris, D.~Berenson, O.~Brock, A.~Causo, K.~Hauser, K.~Okada,
  A.~Rodriguez, J.~M. Romano, and P.~R. Wurman, ``Analysis and observations
  from the first amazon picking challenge,'' \emph{IEEE Transactions on
  Automation Science and Engineering}, vol.~15, no.~1, pp. 172--188, 2016.

\bibitem{buehler2009darpa}
M.~Buehler, K.~Iagnemma, and S.~Singh, \emph{The DARPA urban challenge:
  autonomous vehicles in city traffic}.\hskip 1em plus 0.5em minus 0.4em\relax
  Springer, 2009, vol.~56.

\bibitem{krotkov2017darpa}
E.~Krotkov, D.~Hackett, L.~Jackel, M.~Perschbacher, J.~Pippine, J.~Strauss,
  G.~Pratt, and C.~Orlowski, ``The darpa robotics challenge finals: Results and
  perspectives,'' \emph{Journal of Field Robotics}, vol.~34, no.~2, pp.
  229--240, 2017.

\bibitem{seetharaman2006unmanned}
G.~Seetharaman, A.~Lakhotia, and E.~P. Blasch, ``Unmanned vehicles come of age:
  The darpa grand challenge,'' \emph{Computer}, vol.~39, no.~12, pp. 26--29,
  2006.

\bibitem{todorov2012mujoco}
E.~Todorov, T.~Erez, and Y.~Tassa, ``Mujoco: A physics engine for model-based
  control,'' in \emph{International Conference on Intelligent Robots and
  Systems}.\hskip 1em plus 0.5em minus 0.4em\relax IEEE, 2012, pp. 5026--5033.

\bibitem{yu2020meta}
T.~Yu, D.~Quillen, Z.~He, R.~Julian, K.~Hausman, C.~Finn, and S.~Levine,
  ``Meta-world: A benchmark and evaluation for multi-task and meta
  reinforcement learning,'' in \emph{Conference on Robot Learning}.\hskip 1em
  plus 0.5em minus 0.4em\relax PMLR, 2020, pp. 1094--1100.

\bibitem{brockman2016openai}
G.~Brockman, V.~Cheung, L.~Pettersson, J.~Schneider, J.~Schulman, J.~Tang, and
  W.~Zaremba, ``Openai gym,'' \emph{arXiv preprint arXiv:1606.01540}, 2016.

\bibitem{zhu2020robosuite}
Y.~Zhu, J.~Wong, A.~Mandlekar, and R.~Mart{\'\i}n-Mart{\'\i}n, ``robosuite: A
  modular simulation framework and benchmark for robot learning,'' \emph{arXiv
  preprint arXiv:2009.12293}, 2020.

\bibitem{dasari2019robonet}
S.~Dasari, F.~Ebert, S.~Tian, S.~Nair, B.~Bucher, K.~Schmeckpeper, S.~Singh,
  S.~Levine, and C.~Finn, ``Robonet: Large-scale multi-robot learning,''
  \emph{arXiv preprint arXiv:1910.11215}, 2019.

\bibitem{mandlekar2018roboturk}
A.~Mandlekar, Y.~Zhu, A.~Garg, J.~Booher, M.~Spero, A.~Tung, J.~Gao, J.~Emmons,
  A.~Gupta, E.~Orbay \emph{et~al.}, ``Roboturk: A crowdsourcing platform for
  robotic skill learning through imitation,'' in \emph{Conference on Robot
  Learning}.\hskip 1em plus 0.5em minus 0.4em\relax PMLR, 2018, pp. 879--893.

\bibitem{collins2019benchmarking}
J.~Collins, J.~McVicar, D.~Wedlock, R.~Brown, D.~Howard, and J.~Leitner,
  ``Benchmarking simulated robotic manipulation through a real world dataset,''
  \emph{IEEE Robotics and Automation Letters}, vol.~5, no.~1, pp. 250--257,
  2019.

\bibitem{dean2022robots}
V.~Dean, Y.~G. Shavit, and A.~Gupta, ``Robots on demand: A democratized
  robotics research cloud,'' in \emph{Conference on Robot Learning}.\hskip 1em
  plus 0.5em minus 0.4em\relax PMLR, 2022, pp. 1769--1775.

\bibitem{pickem2017robotarium}
D.~Pickem, P.~Glotfelter, L.~Wang, M.~Mote, A.~Ames, E.~Feron, and
  M.~Egerstedt, ``The robotarium: A remotely accessible swarm robotics research
  testbed,'' in \emph{International Conference on Robotics and
  Automation}.\hskip 1em plus 0.5em minus 0.4em\relax IEEE, 2017, pp.
  1699--1706.

\bibitem{kumar2019offworld}
A.~Kumar, T.~Buckley, J.~B. Lanier, Q.~Wang, A.~Kavelaars, and I.~Kuzovkin,
  ``Offworld gym: open-access physical robotics environment for real-world
  reinforcement learning benchmark and research,'' \emph{arXiv preprint
  arXiv:1910.08639}, 2019.

\bibitem{sun2021research}
Y.~Sun, J.~Falco, M.~A. Roa, and B.~Calli, ``Research challenges and progress
  in robotic grasping and manipulation competitions,'' \emph{IEEE Robotics and
  Automation Letters}, vol.~7, no.~2, pp. 874--881, 2021.

\bibitem{liu2021ocrtoc}
Z.~Liu, W.~Liu, Y.~Qin, F.~Xiang, M.~Gou, S.~Xin, M.~A. Roa, B.~Calli, H.~Su,
  Y.~Sun \emph{et~al.}, ``Ocrtoc: A cloud-based competition and benchmark for
  robotic grasping and manipulation,'' \emph{IEEE Robotics and Automation
  Letters}, vol.~7, no.~1, pp. 486--493, 2021.

\bibitem{funk2021benchmarking}
N.~Funk, C.~Schaff, R.~Madan, T.~Yoneda, J.~U. De~Jesus, J.~Watson, E.~K.
  Gordon, F.~Widmaier, S.~Bauer, S.~S. Srinivasa \emph{et~al.}, ``Benchmarking
  structured policies and policy optimization for real-world dexterous object
  manipulation,'' \emph{arXiv preprint arXiv:2105.02087}, 2021.

\bibitem{pinto2016supersizing}
L.~Pinto and A.~Gupta, ``Supersizing self-supervision: Learning to grasp from
  50k tries and 700 robot hours,'' in \emph{International Conference on
  Robotics and Automation}.\hskip 1em plus 0.5em minus 0.4em\relax IEEE, 2016,
  pp. 3406--3413.

\bibitem{levine2018learning}
S.~Levine, P.~Pastor, A.~Krizhevsky, J.~Ibarz, and D.~Quillen, ``Learning
  hand-eye coordination for robotic grasping with deep learning and large-scale
  data collection,'' \emph{The International Journal of Robotics Research},
  vol.~37, no. 4-5, pp. 421--436, 2018.

\bibitem{sharma2018multiple}
P.~Sharma, L.~Mohan, L.~Pinto, and A.~Gupta, ``Multiple interactions made easy
  (mime): Large scale demonstrations data for imitation,'' in \emph{Conference
  on Robot Learning}.\hskip 1em plus 0.5em minus 0.4em\relax PMLR, 2018, pp.
  906--915.

\bibitem{bahl2021hierarchical}
S.~Bahl, A.~Gupta, and D.~Pathak, ``Hierarchical neural dynamic policies,''
  \emph{arXiv preprint arXiv:2107.05627}, 2021.

\bibitem{kumar2015mujoco}
V.~Kumar and E.~Todorov, ``Mujoco haptix: A virtual reality system for hand
  manipulation,'' in \emph{International Conference on Humanoid Robots}.\hskip
  1em plus 0.5em minus 0.4em\relax IEEE, 2015, pp. 657--663.

\bibitem{ke2021grasping}
L.~Ke, J.~Wang, T.~Bhattacharjee, B.~Boots, and S.~Srinivasa, ``Grasping with
  chopsticks: Combating covariate shift in model-free imitation learning for
  fine manipulation,'' in \emph{International Conference on Robotics and
  Automation}.\hskip 1em plus 0.5em minus 0.4em\relax IEEE, 2021.

\bibitem{grill2020bootstrap}
J.-B. Grill, F.~Strub, F.~Altch{\'e}, C.~Tallec, P.~Richemond, E.~Buchatskaya,
  C.~Doersch, B.~Avila~Pires, Z.~Guo, M.~Gheshlaghi~Azar \emph{et~al.},
  ``Bootstrap your own latent-a new approach to self-supervised learning,''
  \emph{Advances in Neural Information Processing Systems}, vol.~33, pp.
  21\,271--21\,284, 2020.

\bibitem{he2020momentum}
K.~He, H.~Fan, Y.~Wu, S.~Xie, and R.~Girshick, ``Momentum contrast for
  unsupervised visual representation learning,'' in \emph{Conference on
  Computer Vision and Pattern Recognition}.\hskip 1em plus 0.5em minus
  0.4em\relax IEEE, 2020, pp. 9729--9738.

\bibitem{Parisi2022PVR}
S.~Parisi, A.~Rajeswaran, S.~Purushwalkam, and A.~K. Gupta, ``The unsurprising
  effectiveness of pre-trained vision models for control,'' in \emph{ICML},
  2022.

\bibitem{he2016deep}
K.~He, X.~Zhang, S.~Ren, and J.~Sun, ``Deep residual learning for image
  recognition,'' in \emph{Conference on Computer Vision and Pattern
  Recognition}.\hskip 1em plus 0.5em minus 0.4em\relax IEEE, 2016, pp.
  770--778.

\bibitem{nair2022r3m}
S.~Nair, A.~Rajeswaran, V.~Kumar, C.~Finn, and A.~Gupta, ``R3m: A universal
  visual representation for robot manipulation,'' \emph{arXiv preprint
  arXiv:2203.12601}, 2022.

\bibitem{grauman2022ego4d}
K.~Grauman, A.~Westbury, E.~Byrne, Z.~Chavis, A.~Furnari, R.~Girdhar,
  J.~Hamburger, H.~Jiang, M.~Liu, X.~Liu \emph{et~al.}, ``Ego4d: Around the
  world in 3,000 hours of egocentric video,'' in \emph{Conference on Computer
  Vision and Pattern Recognition}.\hskip 1em plus 0.5em minus 0.4em\relax IEEE,
  2022, pp. 18\,995--19\,012.

\bibitem{radford2021learning}
A.~Radford, J.~W. Kim, C.~Hallacy, A.~Ramesh, G.~Goh, S.~Agarwal, G.~Sastry,
  A.~Askell, P.~Mishkin, J.~Clark \emph{et~al.}, ``Learning transferable visual
  models from natural language supervision,'' in \emph{International Conference
  on Machine Learning}.\hskip 1em plus 0.5em minus 0.4em\relax PMLR, 2021, pp.
  8748--8763.

\bibitem{chen2021decision}
L.~Chen, K.~Lu, A.~Rajeswaran, K.~Lee, A.~Grover, M.~Laskin, P.~Abbeel,
  A.~Srinivas, and I.~Mordatch, ``Decision transformer: Reinforcement learning
  via sequence modeling,'' \emph{Advances in Neural Information Processing
  Systems}, vol.~34, pp. 15\,084--15\,097, 2021.

\bibitem{pari2021surprising}
J.~Pari, N.~M. Shafiullah, S.~P. Arunachalam, and L.~Pinto, ``The surprising
  effectiveness of representation learning for visual imitation,'' \emph{arXiv
  preprint arXiv:2112.01511}, 2021.

\bibitem{mansimov2018simple}
\BIBentryALTinterwordspacing
E.~Mansimov and K.~Cho, ``Simple nearest neighbor policy method for continuous
  control tasks,'' 2018. [Online]. Available:
  \url{https://openreview.net/forum?id=ByL48G-AW}
\BIBentrySTDinterwordspacing

\bibitem{kostrikov2021offline}
I.~Kostrikov, A.~Nair, and S.~Levine, ``Offline reinforcement learning with
  implicit q-learning,'' \emph{arXiv preprint arXiv:2110.06169}, 2021.

\bibitem{seno2021d3rlpy}
T.~Seno and M.~Imai, ``d3rlpy: An offline deep reinforcement learning
  library,'' \emph{arXiv preprint arXiv:2111.03788}, 2021.

\end{thebibliography}

\newpage
\section{Appendix}

\subsection{Task Specifications}

\begin{figure}[ht]
\centering

\begin{subfigure}[b]{0.48\linewidth}
    \includegraphics[width=\linewidth]{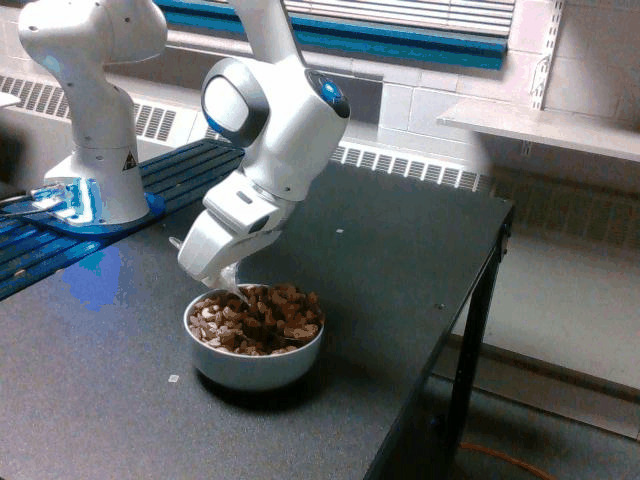}
\end{subfigure}
\begin{subfigure}[b]{0.48\linewidth}
    \includegraphics[width=\linewidth]{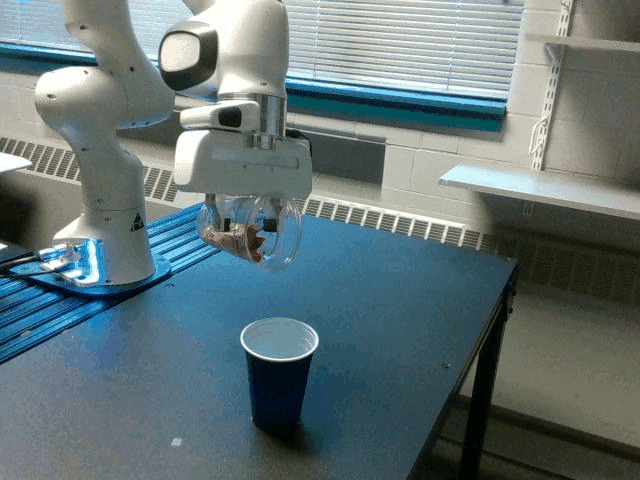}
\end{subfigure}
\begin{subfigure}[b]{0.48\linewidth}
    \vspace{0.3em}
    \includegraphics[width=\linewidth]{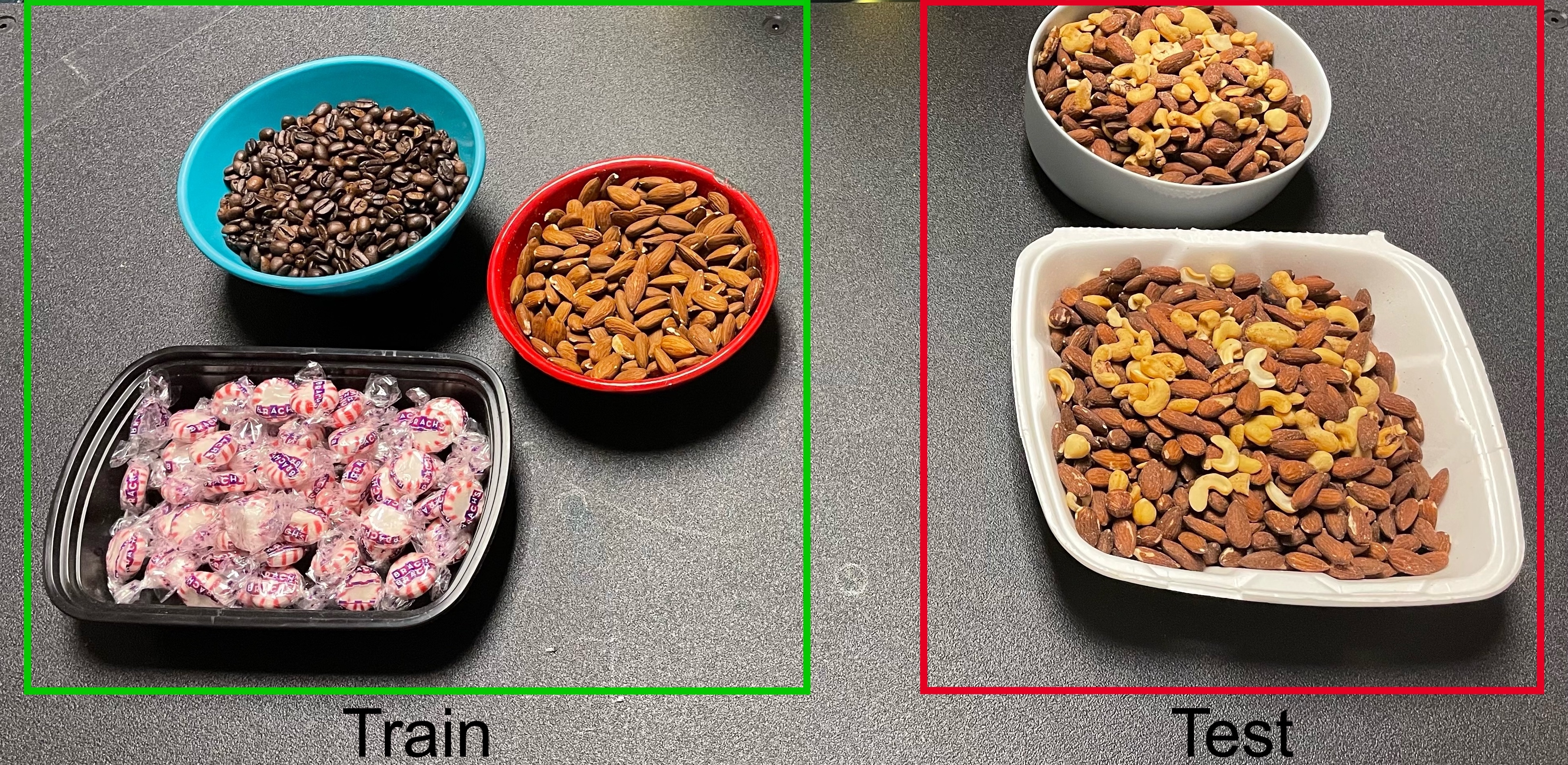}
\end{subfigure}
\begin{subfigure}[b]{0.48\linewidth}
    \vspace{0.3em}
    \includegraphics[trim=0 5mm 0 10mm, clip, width=\linewidth]{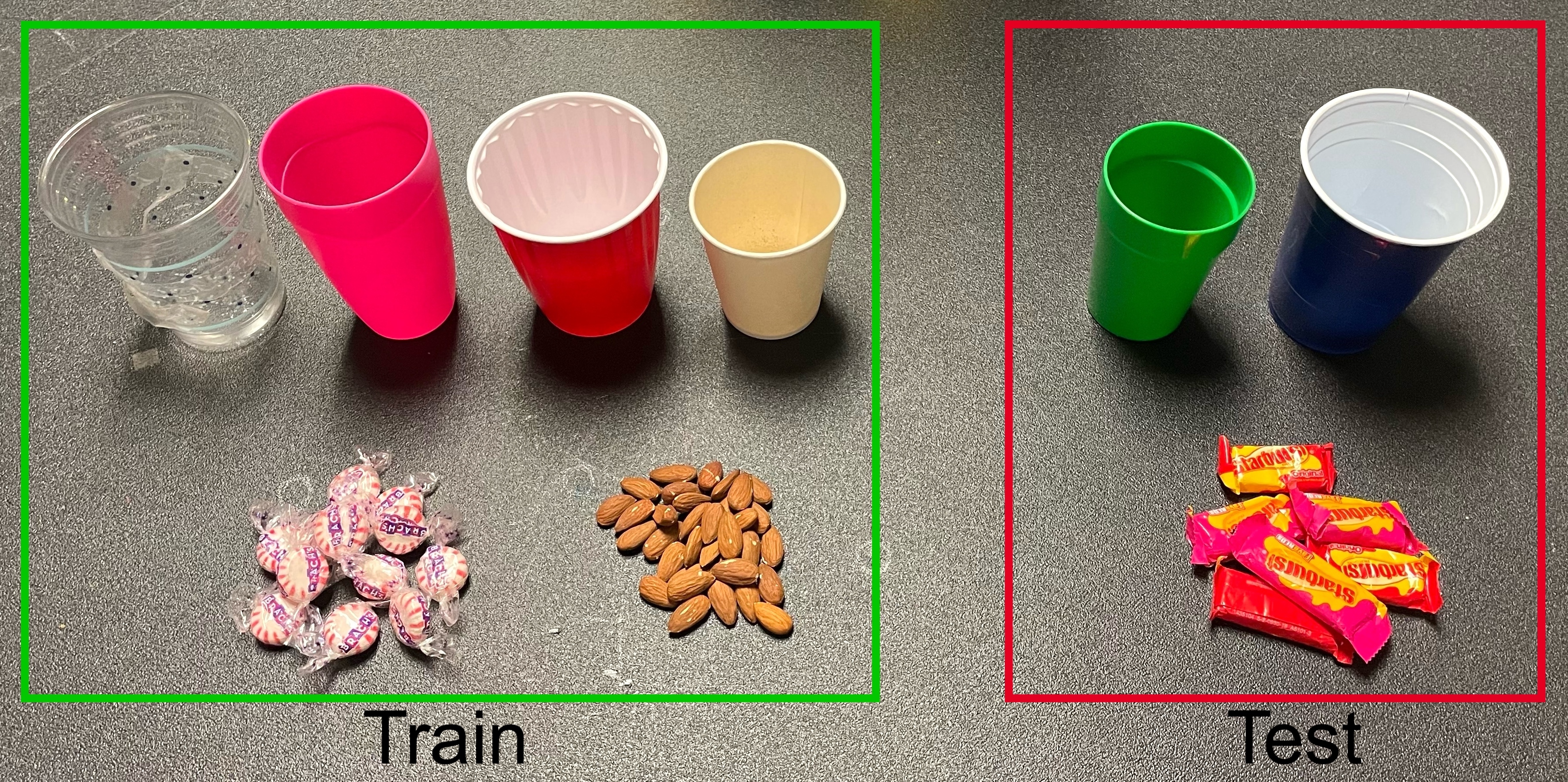}
\end{subfigure}

\caption{\textbf{Benchmark Suite.} Our benchmark includes two tasks: scooping (left) and pouring (right). The bottom images display the train and test objects for each task, respectively.}
\vspace{-0.25cm}
\label{fig:objects}
\end{figure}

Our benchmark contains two manipulation tasks: scooping and pouring. Fig. \ref{fig:objects} shows the train and test objects for each task. The containers have varied sizes, shapes, materials, and colors. The materials being scooped or poured have diverse appearances, granularities, and densities. This  variation enables us to evaluate the generalization capabilities of both visual representations and learned policies.

\begin{figure}[ht]
\centering

    \includegraphics[width=0.9\linewidth]{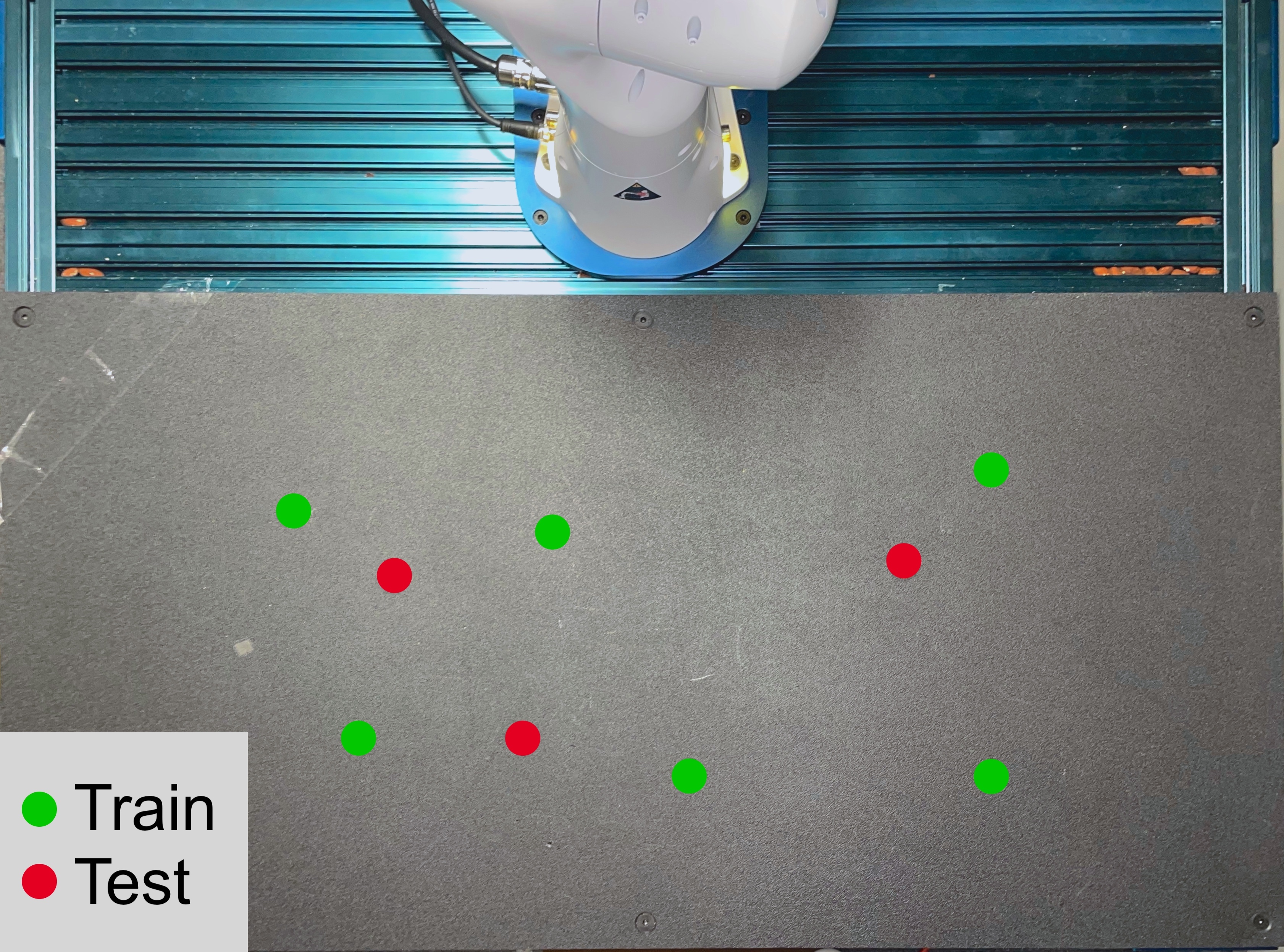}

\caption{\textbf{Train and Test Locations.} The green locations are used in the training data. The red locations are only used at test time.}
\vspace{-0.25cm}
\label{fig:locations}
\end{figure}

Fig. \ref{fig:locations} shows the train and test locations for the center of the container in both tasks. These locations are distributed across a workspace measuring 60$cm$ x 110$cm$. \TODO{I measured this myself, is there better size info for this?} To ensure evaluation consistency, we have marked each location for future reference.

\subsection{Dataset}

\begin{figure}[ht]
\centering

\begin{subfigure}[b]{0.48\linewidth}
    \includegraphics[width=\linewidth]{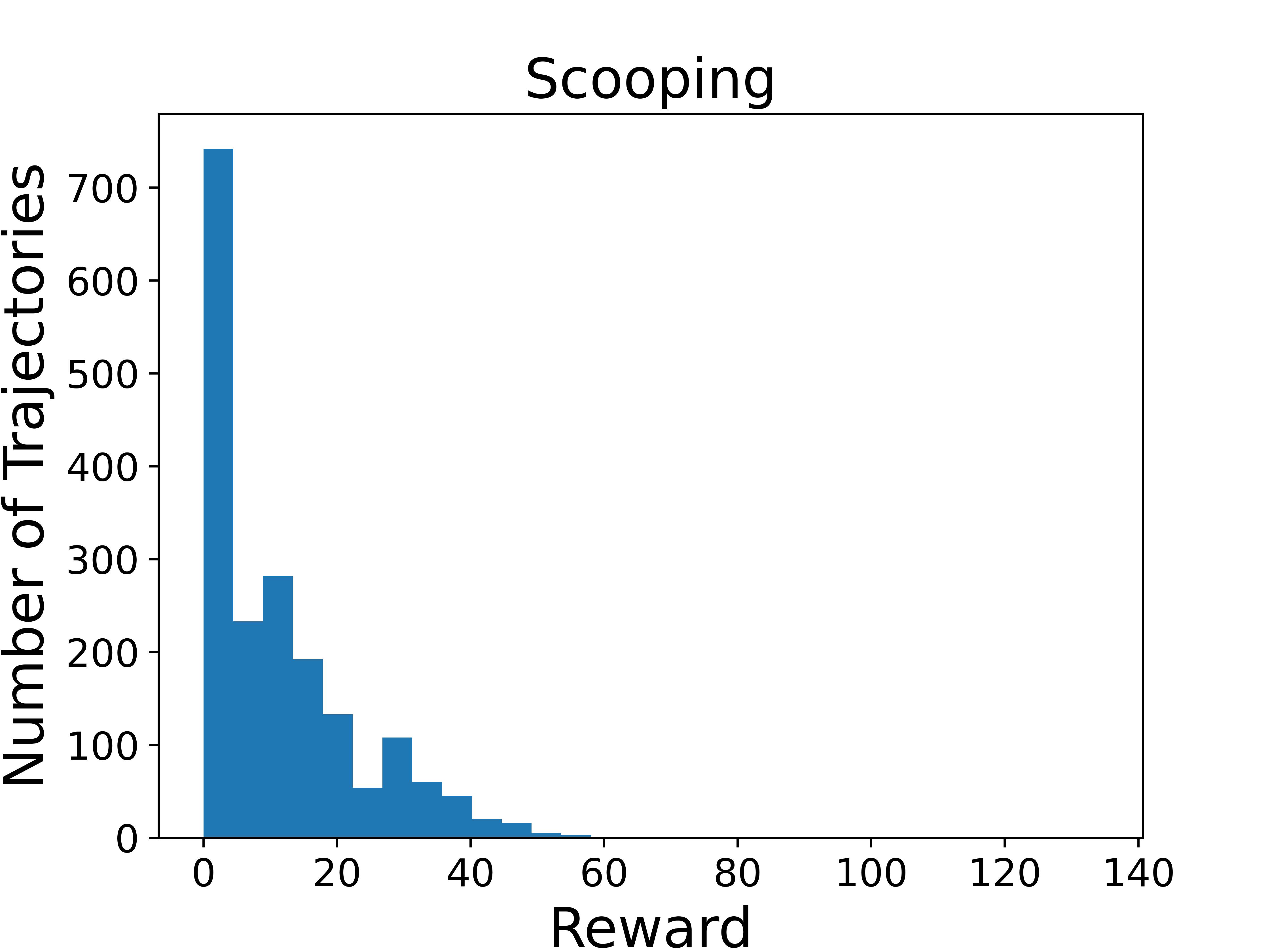}
\end{subfigure}
\begin{subfigure}[b]{0.48\linewidth}
    \includegraphics[width=\linewidth]{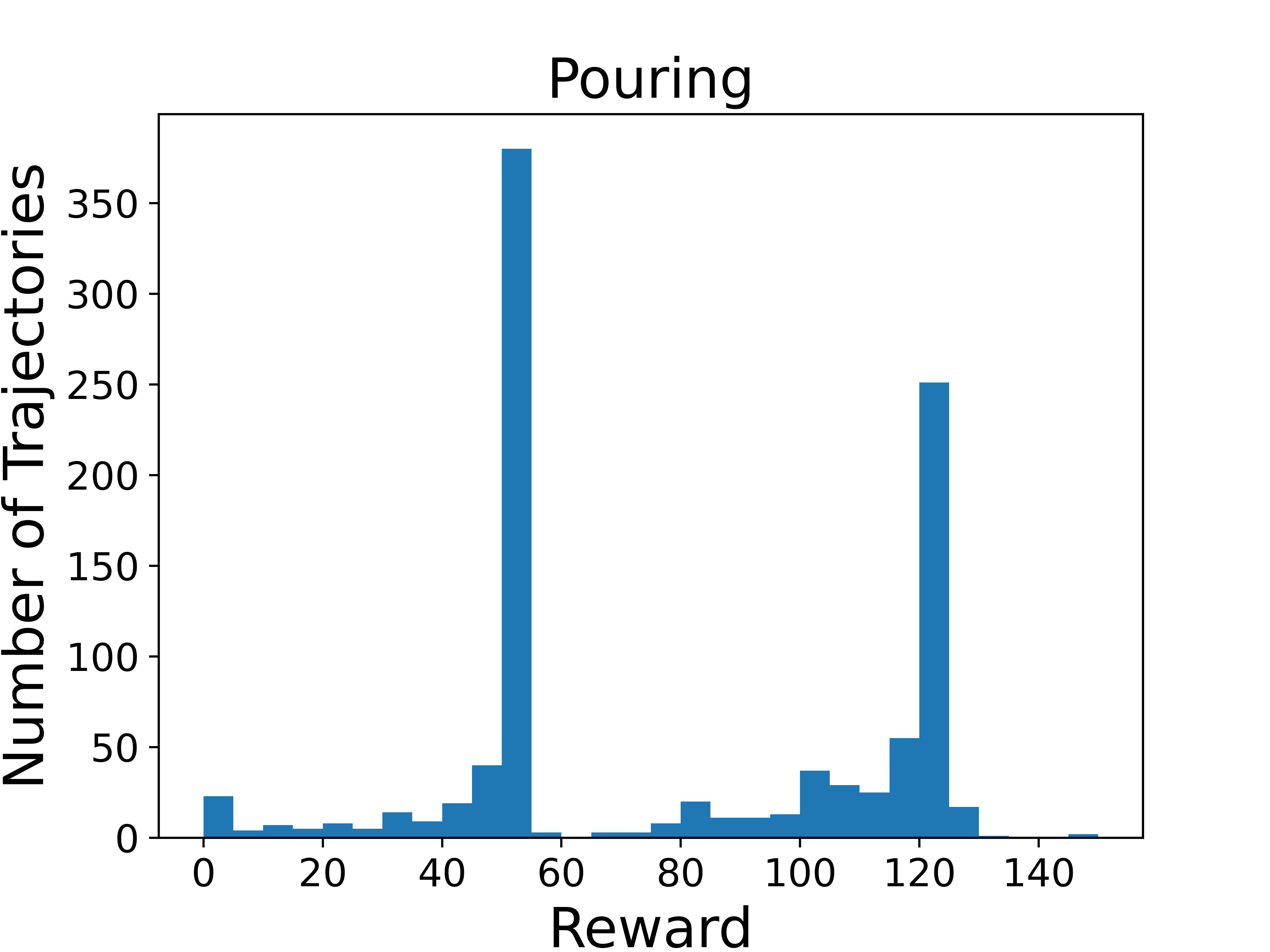}
\end{subfigure}

\caption{\textbf{Reward Distribution.} Each histogram depicts the task rewards in the training dataset. }
\vspace{-0.25cm}
\label{fig:reward_count}
\end{figure}

We compute the reward distribution of our dataset for both tasks (Fig. \ref{fig:reward_count}). In the pouring task, the presence of multiple peaks in the distribution is attributed to the weight differences of the materials poured during training.

We note that the reward values depicted in Fig. \ref{fig:reward_count} are not normalized. Instead, they represent the weight in grams of the material successfully scooped or poured. This unnormalized representation provides a direct measure of the task performance in terms of the weight of the manipulated material.

\subsection{Get Started with TOTO}

We are committed to maintaining the TOTO setup in the long term and continuously seeking additional baselines. We warmly invite the academic and research community to participate in the TOTO benchmark challenges, which include:
\begin{itemize}
   \item \textbf{Visual Representation Model Challenge}:  In this challenge, participants are encouraged to train and submit Behavior Cloning (BC) agents that utilize a pre-trained visual representation model.
    \item \textbf{Agent Policy Challenge}: Participants can choose to either develop a custom visual representation model or utilize the visual representations provided by TOTO. The challenge focuses on designing and submitting agent policies that demonstrate effective manipulation skills.
\end{itemize}
For comprehensive instructions on how to get started, as well as to access our code and dataset, we encourage you to visit our website at  \url{https://toto-benchmark.org/}.

\end{document}